\documentclass[10pt,twocolumn,letterpaper]{article}

\usepackage{wacv}      

\usepackage{graphicx}
\usepackage{amsmath}
\usepackage{amssymb}
\usepackage{booktabs}
\usepackage{multirow}
\usepackage{siunitx}
\usepackage{bbm}
\usepackage{svg}
\usepackage{multirow}

\usepackage[pagebackref,breaklinks,colorlinks]{hyperref}

\usepackage[capitalize]{cleveref}
\crefname{section}{Sec.}{Secs.}
\Crefname{section}{Section}{Sections}
\Crefname{table}{Table}{Tables}
\crefname{table}{Tab.}{Tabs.}

\def\wacvPaperID{2} 
\def\confName{WACV}
\def\confYear{2025}

\begin{document}

\title{Scenario Understanding of Traffic Scenes Through Large Visual Language Models}

\author{Esteban Rivera, Jannik Lübberstedt, Nico Uhlemann, Markus Lienkamp\\
Technical University of Munich, Germany; School of Engineering \& Design, Institute\\
of Automotive Technology and Munich Institute of Robotics and Machine Intelligence\\
{\tt\small esteban.rivera@tum.de}
}
\maketitle

\begin{abstract}
Deep learning models for autonomous driving, encompassing perception, planning, and control, depend on vast datasets to achieve their high performance. However, their generalization often suffers due to domain-specific data distributions, making an effective scene-based categorization of samples necessary to improve their reliability across diverse domains. Manual captioning, though valuable, is both labor-intensive and time-consuming, creating a bottleneck in the data annotation process. Large Visual Language Models (LVLMs) present a compelling solution by automating image analysis and categorization through contextual queries, often without requiring retraining for new categories. In this study, we evaluate the capabilities of LVLMs, including GPT-4 and LLaVA, to understand and classify urban traffic scenes on both an in-house dataset and the BDD100K. We propose a scalable captioning pipeline that integrates state-of-the-art models, enabling a flexible deployment on new datasets. Our analysis, combining quantitative metrics with qualitative insights, demonstrates the effectiveness of LVLMs to understand urban traffic scenarios and highlights their potential as an efficient tool for data-driven advancements in autonomous driving.
\end{abstract}

\section{Introduction}
Models developed for autonomous driving require large amounts of training data to achieve acceptable functionality. However, even with large datasets, their performance can only be optimal within specific operational domains given the underlying data distribution \cite{deepti_hegde__2024}. Hence, to improve their generalization capabilities and, thus, model performance, it is essential to categorize the data to identify which situations are contained within the data distribution. Through this characterization, models can be optimized for a broader range of real-world scenarios, improving their overall reliability and robustness. In the case of autonomous driving, the categorization is performed through tags highlighting the properties of a scene \cite{yu2020bdd100kdiversedrivingdataset}. As a prominent example, the daytime of a scene is important as camera object detectors trained solely on samples with sunlight suffer from decreased scores for nighttime scenes \cite{LI2021102946}. Similarly, crowded scenes provide challenges for different driving modules if only scenarios containing a few agents were present during training \cite{9659743}. These two cases illustrate the need to choose suitable tags based on which an adequate distribution can be formed. Not only does this allow for a more balanced selection and subsequent training, but it also introduces a more effective workflow by replacing manual sample selection with a simple definition of the desired categories. In addition, it saves time and money as the manual labeling process can be both, costly and cumbersome.

\begin{figure}[]
    \centering
    \includegraphics[width=0.9\linewidth]{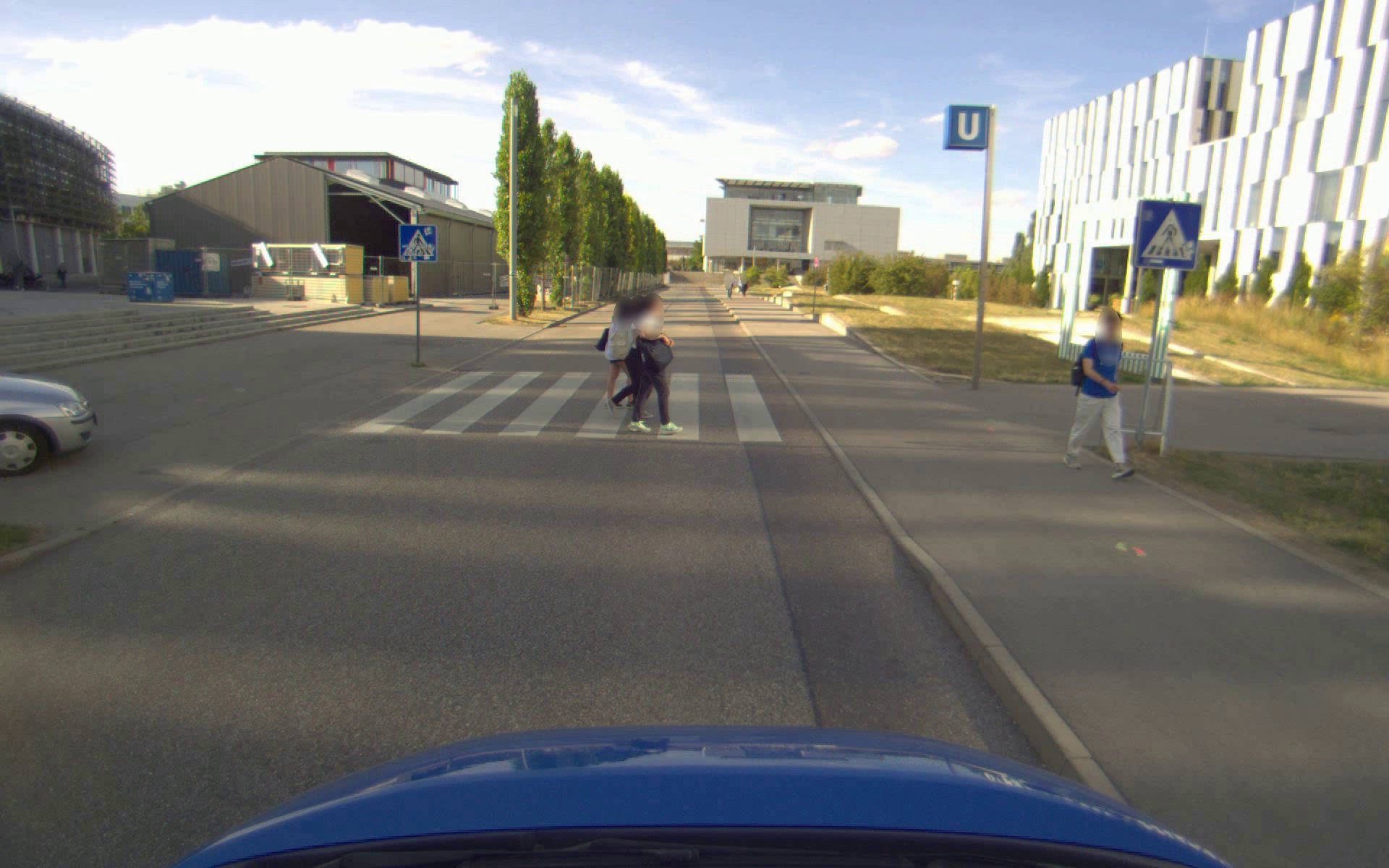}
    \caption{Example for a common traffic scenario contained in our dataset, depicting the interaction of the ego vehicle with a group of pedestrians crossing over a crosswalk on a sunny day and a silver car parked on the sidewalk towards the left.}
    \label{fig:intro}
\end{figure}

Large Vision Language Models (LVLMs) have emerged as a promising tool to automate the task of categorization. By providing an image of a scene alongside a query reflecting the desired category, LVLMs can effectively estimate the similarity of the scene with the provided tag, yielding the desired categories after a few post-processing steps. Since these models are trained on vast amounts of data, retraining or fine-tuning is usually not required even when introducing new categories. This is in contrast to classical approaches where often convolutional neural networks are employed to classify a given image. Here, re-labeling of the data followed by re-training of the network is necessary anytime a new category is added, which both involves a significant effort and associated cost. Both, proprietary models like GPT-4 \cite{openai2024gpt4technicalreport} and open-source models like LLaVA \cite{liu2023visualinstructiontuning}, have achieved impressive results on visual questioning tasks recently \cite{yu2024mm}. Therefore, their capability to understand and reason about scenarios relevant for autonomous driving will be investigated in this work, paving the way for their potential application as automated categorizers for large datasets. 

In summary, the contributions of this study are threefold:
\begin{itemize}
    \item First, we provide a quantitative analysis of the scenario understanding of several LVLMs by categorizing traffic scenes for different tags on an in-house and the BDD100K dataset.
    \item Second, we conduct a qualitative analysis of their performance for representative categories, highlighting strengths and shortcomings.
    \item Lastly, we implement an automated categorization pipeline based on the investigated models which is applicable to arbitrary datasets focusing on traffic scenes.\footnote{\url{https://github.com/TUMFTM/CatPipe}}  
\end{itemize}

\section{Related work}

\paragraph{Datasets for autonomous driving} 
Datasets are the foundation for every learning-based approach and subsequently their performance evaluation. Since our goal is to use VLMs to classify images representing scenes in the context of autonomous driving, datasets found within the research community can offer a diverse spectrum of categories. One problem with this approach is that each has different tags containing information about the environment or detected objects themselves. In addition, some categories we want to detect are not captured at all. While information about pedestrians, bicyclists, traffic lights, and traffic signs can be obtained from object detection datasets like BDD100k \cite{yu2020bdd100kdiversedrivingdataset}, Waymo \cite{sun2020scalabilityperceptionautonomousdriving}, Apolloscape \cite{huang2018apolloscape}, Zenseact \cite{zenseact2023} and Cityscapes \cite{cordts2016cityscapes}, weather tags are only available in the metadata of the first two. Moreover, since BDD100k provides additional tags for the urban environment, it is therefore chosen for the verification of our results obtained for the different image classification approaches. Closer to the LVLM application for autonomous driving is the MAPLM dataset and benchmark \cite{cao2024maplm}, which collects multimodal information relevant for question-answering.  We do not look for comparison in the natural-language level from MAPLM, because we want to compare directly the tagging task, which does not require Natural Language capabilities to be solved.

\paragraph{Visual Question Answering} 
The term Visual Question Answering (VQA) describes the general approach of using LVLMs to extract visual information from images constrained by text. This is closely related to LVLM-based image classification. Therefore, we selected the models for the evaluation based on their performance on selected VQA datasets. VQA 2.0 \cite{goyal2017vqa} focuses on general vocabulary without requiring background information about the topics not conveyed in the image. In contrast, Pope \cite{li2023evaluatingobjecthallucinationlarge} is a method to evaluate how susceptive a model is to hallucinations. Its focus is object detection, as it uses the ground truth of the COCO dataset to construct questions. For each object present in an image, positive samples are created. Negative samples are constructed by randomly sampling objects from an object list that are absent from a scene's ground truth. Additionally, we use the MMBench \cite{liu2024mmbenchmultimodalmodelallaround} dataset as it evaluates a model's susceptiveness to circular shift. This is important to reduce the influence of the input order in which the possible categories are provided in. 

\paragraph{Vision Language Models}
The models used for Visual Question Answering are typically LLMs with a vision encoder component. LLaVA uses a model from the CLIP \cite{radford2021learningtransferablevisualmodels} model family to obtain a feature vector of the input image which is projected and fed into the LLM. This approach can be generalized to larger input image resolutions by splitting the image into patches and feeding each patch into the vision encoder as done with
ComposerHD \cite{zhang2023internlmxcomposervisionlanguagelargemodel}. The features per patch are then concatenated before being fed into the LLM. CogVLM \cite{hong2023cogagentvisuallanguagemodel} adds a module called Visual Expert to each attention and fully connected layer to enhance the alignment of visual and language features, essentially doubling the parameters in these components. In autonomous driving, LLMs can be used to obtain a natural language representation of the scene, recommend an action for the driver, and explain its reasoning, as presented by Wayve with the LingoQA dataset \cite{marcu2024lingoqavideoquestionanswering}. Jain et al.\cite{jain2024} compares the performance of GPT4 and LlaVA for the VQA task specifically in the Autonomous Driving domain, but not with the goal of categorize the scenes but to get a human understandable knowledge of the traffic scenes.

\section{Methodology}

In the following, we will introduce our in-house dataset alongside the selected models and our evaluation procedure. Moreover, we provide insights into our prompt design and the metrics to quantify the overall results.

\subsection{Datasets}

As the foundation of our work, the majority of the experiments are conducted on a manually annotated, in-house dataset collected in Germany, which is currently not publicly available. The dataset was created based on several rides with a camera-equipped vehicle over a span of six months. Hence, it contains diverse traffic as well as weather and daytime conditions as can be seen in \cref{sec:results}. For this evaluation, 257 of roughly 800 scenes with a duration of \SI{20}{\second} each were selected and manually annotated based on the median frame with the tags required for the classification task. The data distribution can be seen in the supplementary material. In addition, to validate the results against an independent and publicly available reference, experiments are conducted on the BDD100k dataset \cite{yu2020bdd100kdiversedrivingdataset}. Here, the classification performance of the selected LVLMs is also compared against traditional closed-vocabulary image classification methods.

\subsection{Categories}
\label{subsec:categories}

\begin{table*}[tb]
  \caption{Proposed category grouping into detection and reasoning tasks, alongside the associated tags for each.
  }
  \label{tab:models}
  \centering
  \begin{tabular}{@{}lp{4cm}p{11cm}@{}}
    \toprule
    Task Name & Category Name & Tags \\ \midrule
    \multirow{5}{*}{Detection} & Person & yes, no \\ 
    & Traffic sign for ego-vehicle & yes, no \\ 
    & Traffic light for ego-vehicle & yes, no \\ 
    & Number of vulnerable road users & none, few, several, many \\ 
    & Lane marks & normal lane marks, crosswalk, bus lane, no lane marks \\ 
    \hline
    \multirow{17}{*}{Reasoning} &Vision impairing brightness (VIB)& yes, no \\
    & Weather & rainy, snowy, clear, overcast, partly cloudy, foggy, undefined \\
    & Time of day & twilight, daytime, nighttime, undefined \\ 
    & Land use & urban area, rural area, suburban area, industrial area, nature \\ 
    & Environment & tunnel, residential area, parking lot, city street, gas station, highway, undefined \\ 
    & Road condition & dry road, wet road, snowy road, icy road, muddy road \\
    & Street configuration & one-way street, two-way street \\
    
    & Number of lanes & 0, 1, 2, 3, 4, 5, 6 \\
    & Traffic scene & free-flowing traffic, congested traffic, traffic accident, construction zone \\
    & Road intersection & yes, no \\
    & Vehicle manoeuvre & moving forward, stopped, turning, lane changing, parking \\ 
  \bottomrule
    \label{tab:categories}
  \end{tabular}
\end{table*}
In this work, 16 categories are chosen to investigate the classification performance in an autonomous driving setting. To achieve a better comparability across categories, we group them according to their primary objective, being detection and reasoning. An overview about the respective categories alongside their associated tags can be found in \cref{tab:categories}. In the first case, the detection categories comprise those which can be answered by assessing the presence of one specific class or object within the provided image. Secondly, the reasoning categories require a deeper understanding of the scene by identifying the presence of several objects and determining their relationships to derive a specific conclusion. Here, LVLMs show the most promise when employed over traditional, learning-based image classifiers. For the tag definition, descriptive terms were preferred over plain numbers in most cases since LLMs have showcased weaknesses for these types of tokens in the past \cite{rahmanzadehgervi2024visionlanguagemodelsblind}. 

\subsection{Models}
The models chosen for this study are primarily selected based on their reported performance on the commonly employed datasets VQA 2.0, Pope and MMBench, as well as their code availability. These models are summarized in \cref{tab:benchmarkmodels}, showcasing their reported results on the different datasets mentioned. In addition to comparing different model architectures, we examine variations of LLaVA \cite{liu2023visualinstructiontuning} to determine whether the slight performance gains observed for larger model sizes directly translate to our domain. We also evaluate InternLM-XComposer2-4KHD \cite{zhang2023internlmxcomposervisionlanguagelargemodel} (ComposerHD) using images at resolutions of 336px and 720px to assess the impact of image quality on accuracy. All models, except GPT-4 \cite{openai2024gpt4technicalreport}, are open-source and were evaluated without fine-tuning to assess their out-of-the-box scenario understanding. This approach ensures accessibility for users with limited data or hardware resources while providing insights into the models’ generalizability across diverse urban traffic settings.

\subsection{Prompt engineering}
\begin{figure}[]
    \centering
    \includegraphics[width=\linewidth]{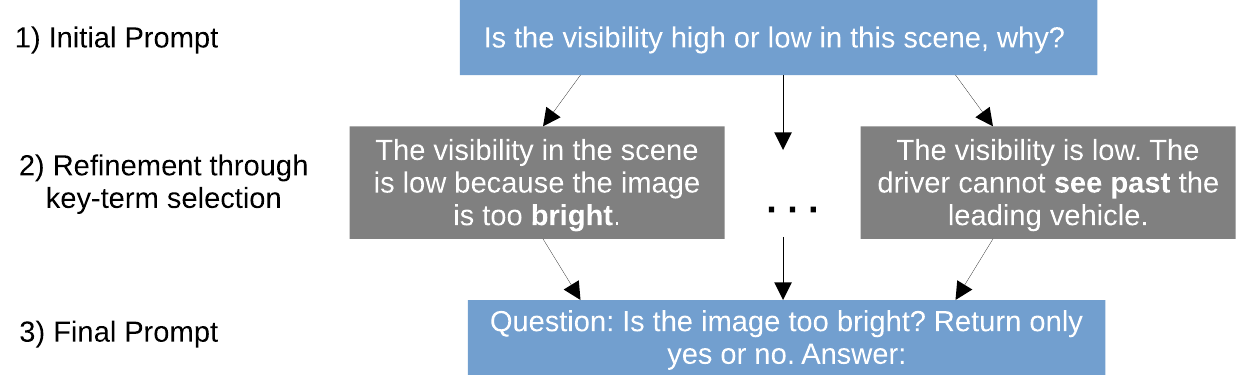}
    \caption{Workflow of our three-step prompt engineering approach based on the example of vision imparing brightness.}
    \label{fig:prompt}
\end{figure}

To design a prompt for our classification task that both generalizes well across the chosen models and increases their corresponding scores, the tags for each category need to be considered. In total, we evaluated three distinct prompt templates of which only the most promising one is showcased in this work. To guide the prompt construction process, we initially create a set of arbitrary prompts describing the classification task. While some of these were manually crafted, we also query a random model\footnote{\href{https://chat.lmsys.org/}{https://chat.lmsys.org/}} to provide a template that can be used for classifications with LLMs, improving the diversity of these initial prompts. Afterwards, as seen \cref{fig:prompt}, the prompts were extended by the expression \textit{", why?"} such that the LLM's reasoning behind the decision is provided alongside its prediction, allowing us to identify common response patterns. Employing five distinct samples, this process was repeated for every category and model showcased in this work, guiding the selection of specific key terms used to construct an more optimized prompt. Since some models fine-tuned for VQA appended \textit{"Question: "} as prefix and \textit{"Answer: "} as suffix to the original prompt template \cite{li2022blipbootstrappinglanguageimagepretraining}, we additionally used this structure after the refinement process, leading us to the final prompts used in this work, which can be seen in the supplementary material.

\subsection{Performance measures}
The performance of the models is measured by the accuracy and macro F1-score \cite{opitz2021macrof1macrof1}. Accuracy is used as it is intuitive and balances out debatable misclassifications for rare tags. The macro F1-score provides a better measure for the class-specific performance which is especially useful for unbalanced classes per category.

\section{Results}
\label{sec:results}
\begin{figure*}[t]
    \centering
    \includesvg[width=\textwidth]{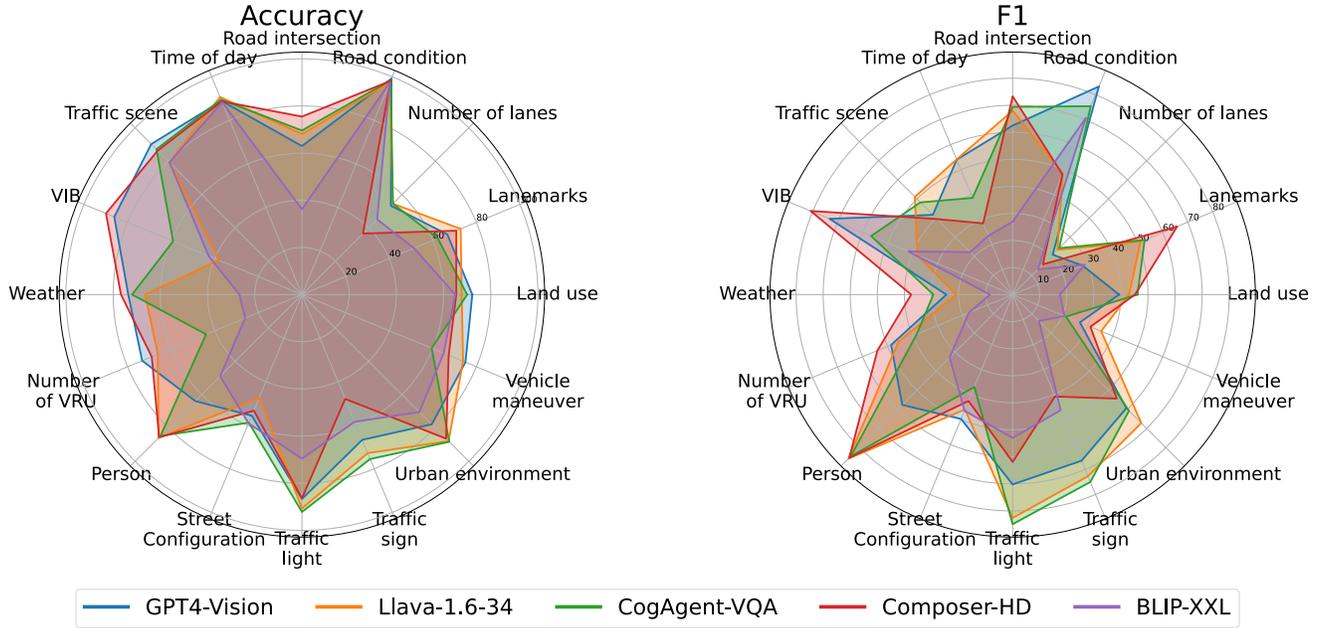}
    \caption{Accuracy and F1-score for the best model of each architecture}
    \label{fig:spider}
\end{figure*}

To illustrate the scores of the best-performing models across various categories, \cref{fig:spider} displays their accuracy and F1 score using two spider plots. Furthermore, to provide a more detailed comparison of the top results for each category introduced in \cref{subsec:categories}, \cref{tab:bestf1} highlights the best models for each category with respect to their F1-score as it is more meaningful for pronounced class-imbalances. Averaging the scores across all categories, GPT4-Vision achieves the highest mean accuracy at 0.75, while CogAgent attains the highest mean F1-score of 0.53. However, this performance varies significantly for individual categories, particularly in cases where some contain only a few samples per tag. Hence, a detailed analysis of these differences is provided in the following section. 

\begin{table}[h]
  \caption{Accuracy of various LVLMs on the VQA datasets VQA 2.0, Pope and MMBench (MMB). GPT-4 did not perform a benchmark specific training.
  }
  \label{tab:benchmarkmodels}
  \centering
  \begin{tabular}{@{}lccc@{}}
    \toprule
    Model (Parameters) & VQA 2.0 & Pope & MMB\\
    \midrule
    GPT4-Vision\cite{openai2024gpt4technicalreport}&77.2&&\\
    LLaVA-1.6-34 (34B)\cite{liu2023visualinstructiontuning} & \textbf{83.7}&87.7&79.3\\
    LLaVA-1.6-13 (13B) & 82.8&86.2&70.0\\
    LLaVA-1.6-7 (7B)&81.8&86.7&68.7\\
    LLaVA-1.5 (13B) & 80.0&85.9&67.8\\
    CogVLM\cite{wang2024cogvlmvisualexpertpretrained} (18B) &83.4&&\\
    CogAgent-VQA\cite{hong2023cogagentvisuallanguagemodel} (18B) &\textbf{83.7}&85.9&\\
    Composer-HD\cite{zhang2023internlmxcomposervisionlanguagelargemodel} (8B) & &&\textbf{80.2}\\
    Deepseek\cite{lu2024deepseekvlrealworldvisionlanguageunderstanding} (8B) & &\textbf{88.1}&73.2\\
    InstructBLIP\cite{dai2023instructblipgeneralpurposevisionlanguagemodels} (13B) & &83.75&\\
    BLIP\cite{li2023blip2bootstrappinglanguageimagepretraining} (8B) & 65.25& &\\
  \bottomrule
  \end{tabular}
\end{table}

\begin{table}[hb!]
\centering
\begin{tabular}{llrr}
\toprule
Category &            Model &  Acc &  F1 \\
\midrule
Land use                        &     Llava-1.6-7(7B) &      66.5 &      51.9 \\
Lanemarks                       &      Composer-HD(8B) &      70.8 &      65.8 \\
Number of lanes                 &     Llava-1.6-7(7B) &      49.4 &      28.6 \\
Number of VRU                   &      Composer-HD(8B) &      68.9 &      54.1 \\
Person                          &     Llava-1.6-13(13B)  &      90.7 &      90.6 \\
Road condition                  &           CogVLM(18B) &      99.2 &      86.6 \\
Road intersection               &      Composer-HD(8B) &     75.5 &      73.3 \\
Street configuration            &         Deepseek(8B) &      63.0 &      57.0 \\
Time of day                     &      Llava-1.6-13(13B) &      90.7 &      54.1 \\
Traffic light                   &         CogAgent(18B) &      92.6 &      86.6 \\
Traffic scene                   &     Llava-1.6-13(13B) &      83.7 &      52.2 \\
Traffic sign                    &     Llava-1.6-7(7B)  &      77.8 &      77.5 \\
Urban environment               &        Llava-1.5(13B) &      89.1 &      67.9 \\
Vehicle maneuver                &     Llava-1.6-34(34B) &      73.9 &      35.4 \\
VIB                             &  Composer-HD-336 &      88.7 &      81.6\\
Weather                         &      Composer-HD(8B) &      76.7 &      37.5 \\
\bottomrule
\end{tabular}
\caption{Investigated categories alongside the metrics of the best-performing model for each case, determined by the F1 score.}
\label{tab:bestf1}
\end{table}

\subsection{Category results}

\begin{figure}[h]
    \centering
    \begin{subfigure}[b]{0.49\linewidth}
        \centering
        \includegraphics[width=\linewidth]{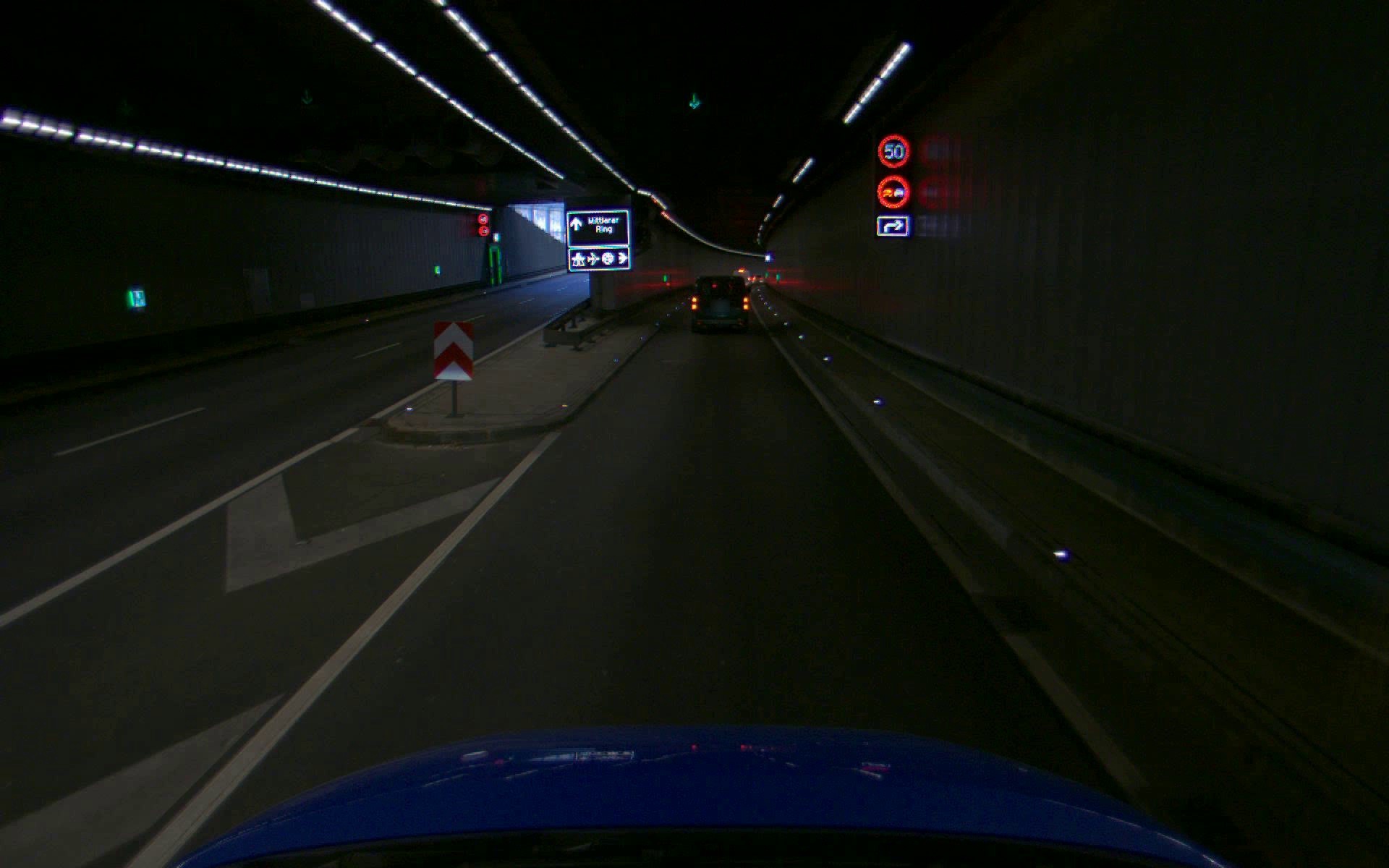}
    \end{subfigure}
    \begin{subfigure}[b]{0.49\linewidth}
        \centering
        \includegraphics[width=\linewidth]{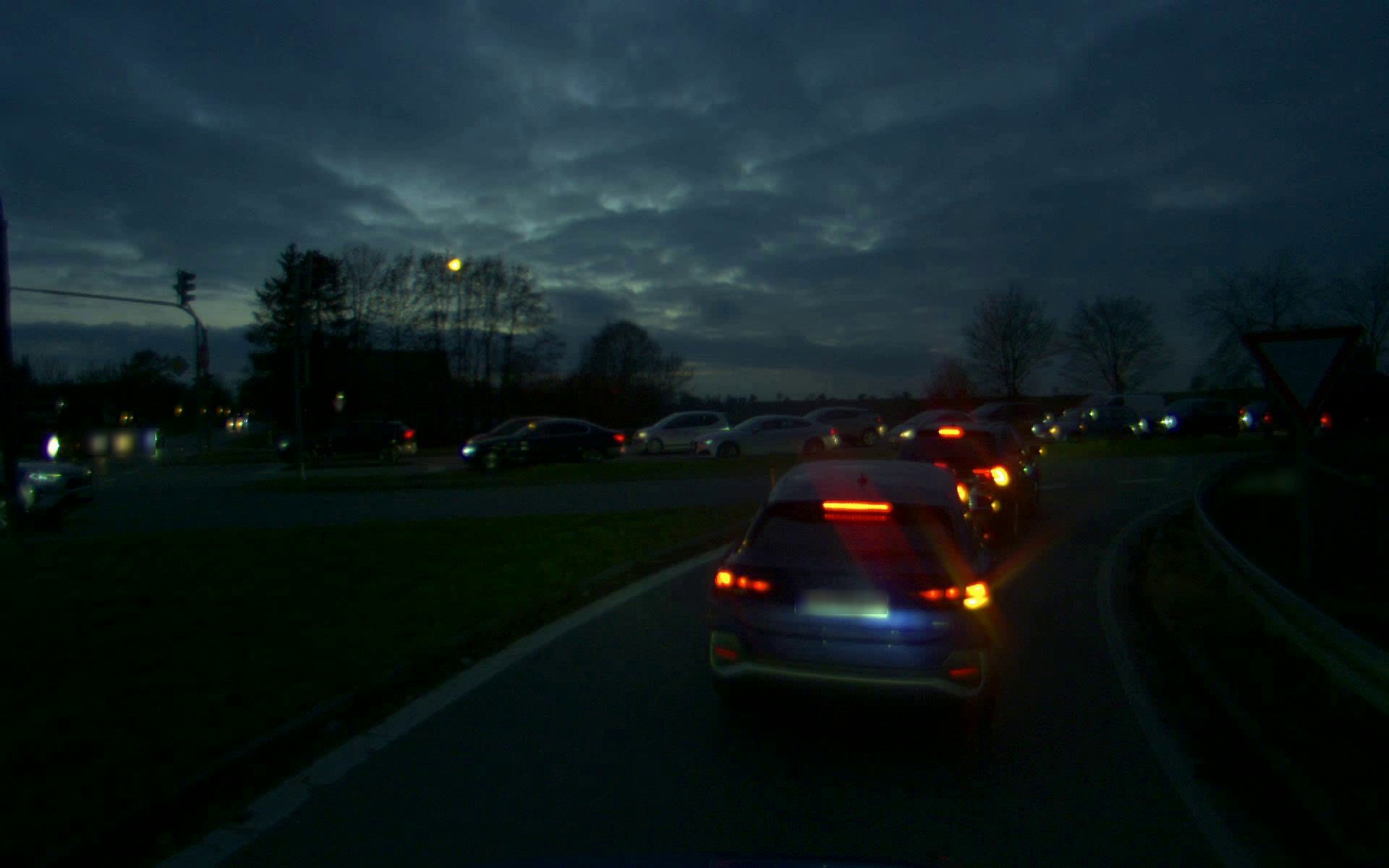}
    \end{subfigure}
    \caption{Examples for false positives in the traffic light category. Here, Composer-HD mistakes illuminated traffic signs (left) or red rear lights of vehicles (right) with actual traffic lights.}
    \label{fig:traffic_light}
\end{figure}

\paragraph{Traffic Lights}
In this category, CogAgent achieves the highest scores, with an accuracy of \SI{93.4}{\percent} and an F1 score of \SI{87.8}{\percent}. As shown on the left side of \cref{fig:spider}, Composer-HD and Llava-1.6-34 perform equally well in terms of accuracy. However, this changes when considering the F1 score, as illustrated on the right, where more pronounced performance gaps emerge. While the models generally perform well under daylight conditions, false-positive predictions of traffic lights in dark conditions significantly impact their performance. This issue arises primarily due to the models' shortcomings to distiguish traffic lights from red rear lights of vehicles or illuminated traffic signs as shown in \cref{fig:traffic_light}.

\paragraph{Lanemarks}
When differenciating different types of lanemarks, Composer-HD achieves the highest F1 score of \SI{66.0}{\percent} and an accuracy with \SI{71.2}{\percent} and. Similar to the traffic light detection, Llava-1.6-34 performs similarly well, while CogAgent now falls behind GPT4-Vision with regards to accuracy. Given the strong class-imbalance for this task as the majority of lanes are white (normal), this order drastically changes for the F1 score, where Composer-HD is over \SI{10}{\percent} ahead of the next best model, CogAgent. During our analysis, the main weaknesses arise from the \textit{no lanemark}-label in scenes with difficult lighting conditions, and the \textit{crosswalk}-label when no crosswalk lines are visible, but pedestrians are crossing the road. Occasionally, a row of snow on the side walk is also mistakenly perceived as a lanemark by most models.

\begin{figure}[h!]
    \centering
    \begin{subfigure}[b]{0.49\linewidth}
        \centering
        \includegraphics[width=\linewidth]{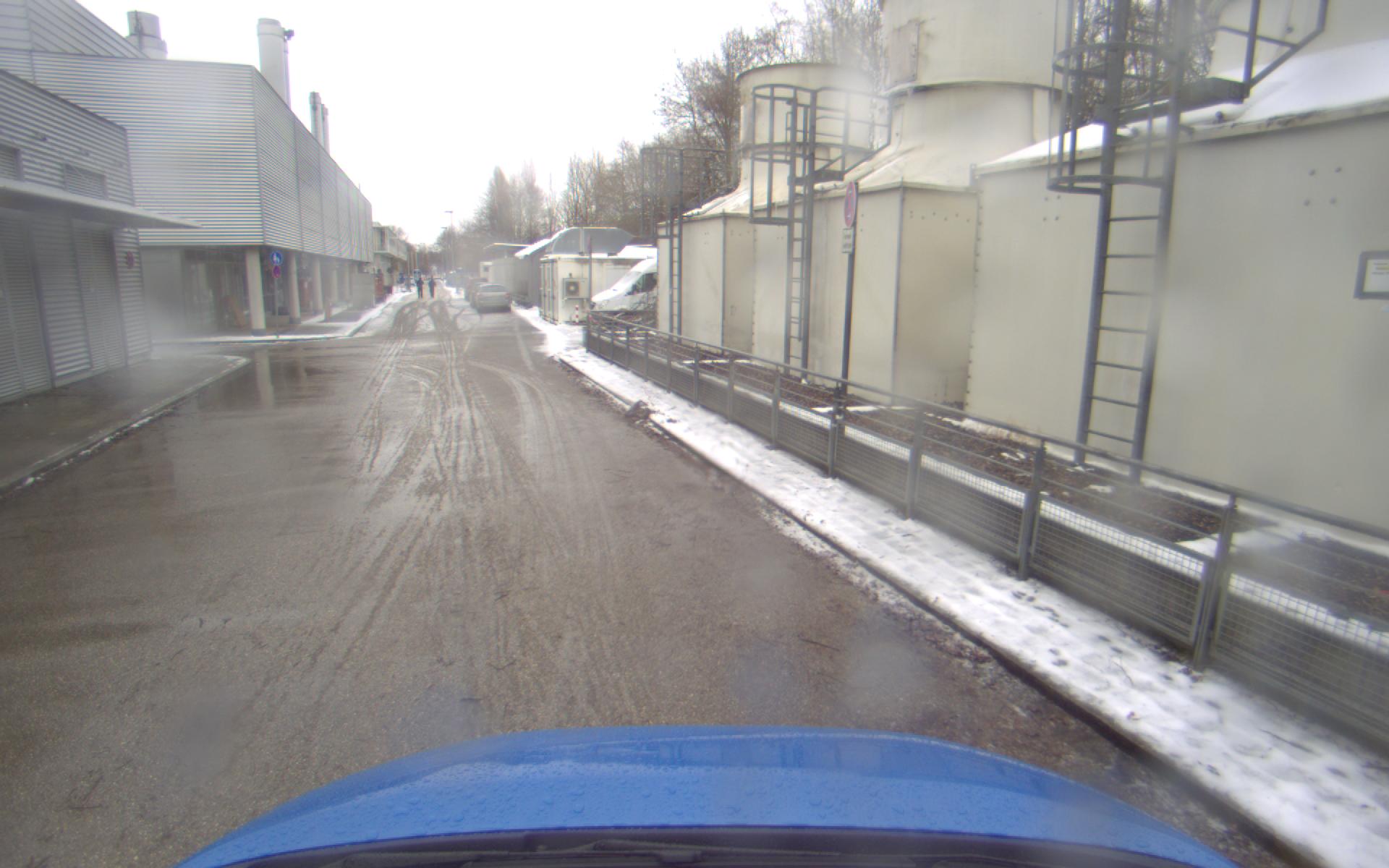}
    \end{subfigure}
    \begin{subfigure}[b]{0.49\linewidth}
        \centering
        \includegraphics[width=\linewidth]{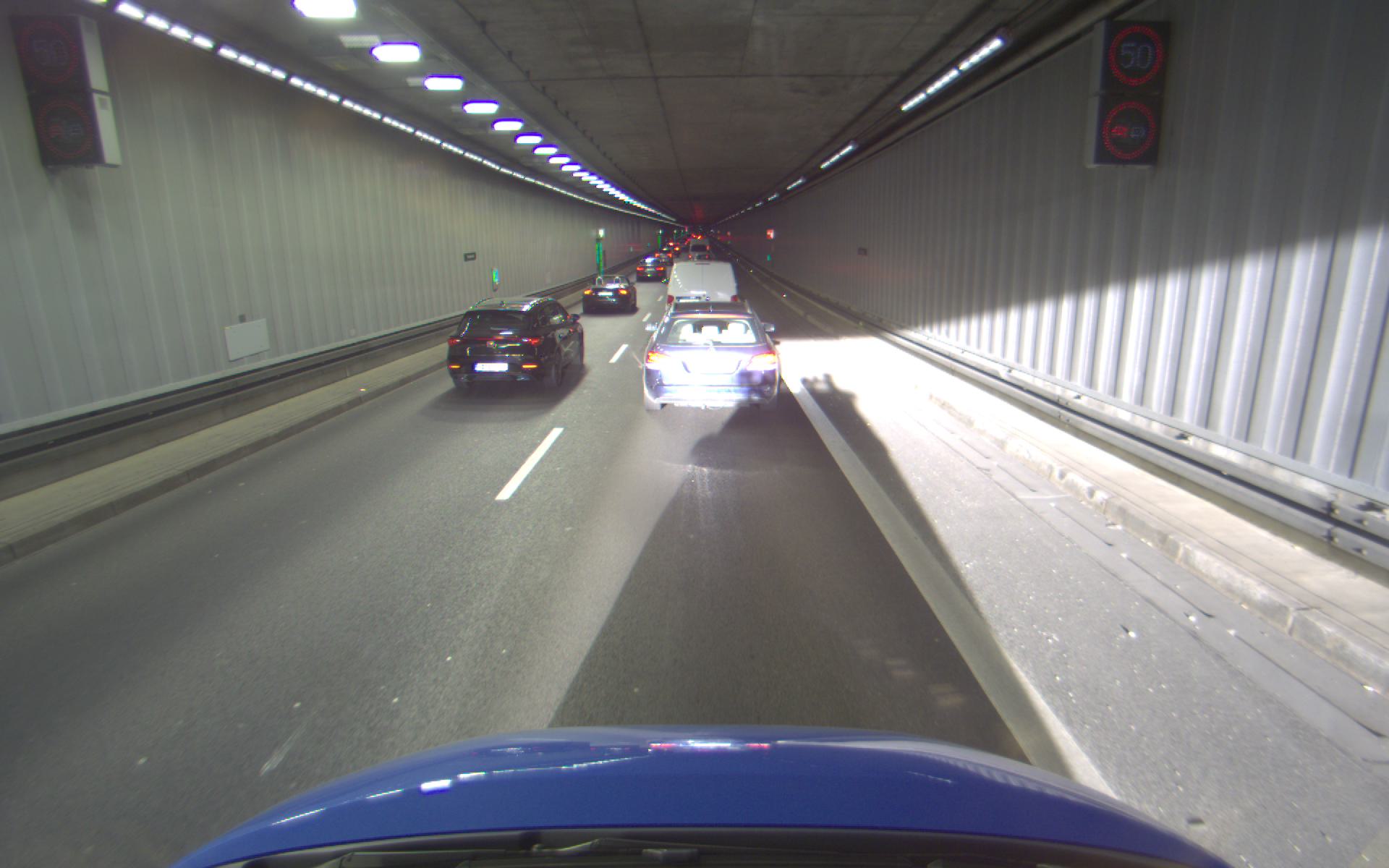}
    \end{subfigure}
    \caption{Example scenarios where the actualy weather condition is difficult to determine. }
    \label{fig:weather}
\end{figure}

\paragraph{Weather}
Classifying weather conditions with a single term is often ambiguous, as multiple tags can apply in many cases. This is evident in \cref{tab:bestf1}, where Composer-HD achieves a high accuracy of \SI{76.7}{\percent}, but its best F1 score is only \SI{37.5}{\percent}. Similarly, looking on the right side of \cref{fig:spider}, all models perform poorly in this category, with most scores ranging from \SIrange{20}{30}{\percent}. The challenges here are illustrated in \cref{fig:weather}, showing a \textit{rainy} and an \textit{undefined} scene. In the first image, most models classified the scene as \textit{snowy} due to visible snow, overlooking the water droplets on the lens indicating rain. In the second image, the tunnel obscures direct weather inference. Nevertheless, most models indicate \textit{clear} conditions which is likely influenced by sunlight shining into the tunnel from behind.

\begin{figure}[h!]
    \centering
    \begin{subfigure}[b]{0.49\linewidth}
        \centering
        \includegraphics[width=\linewidth]{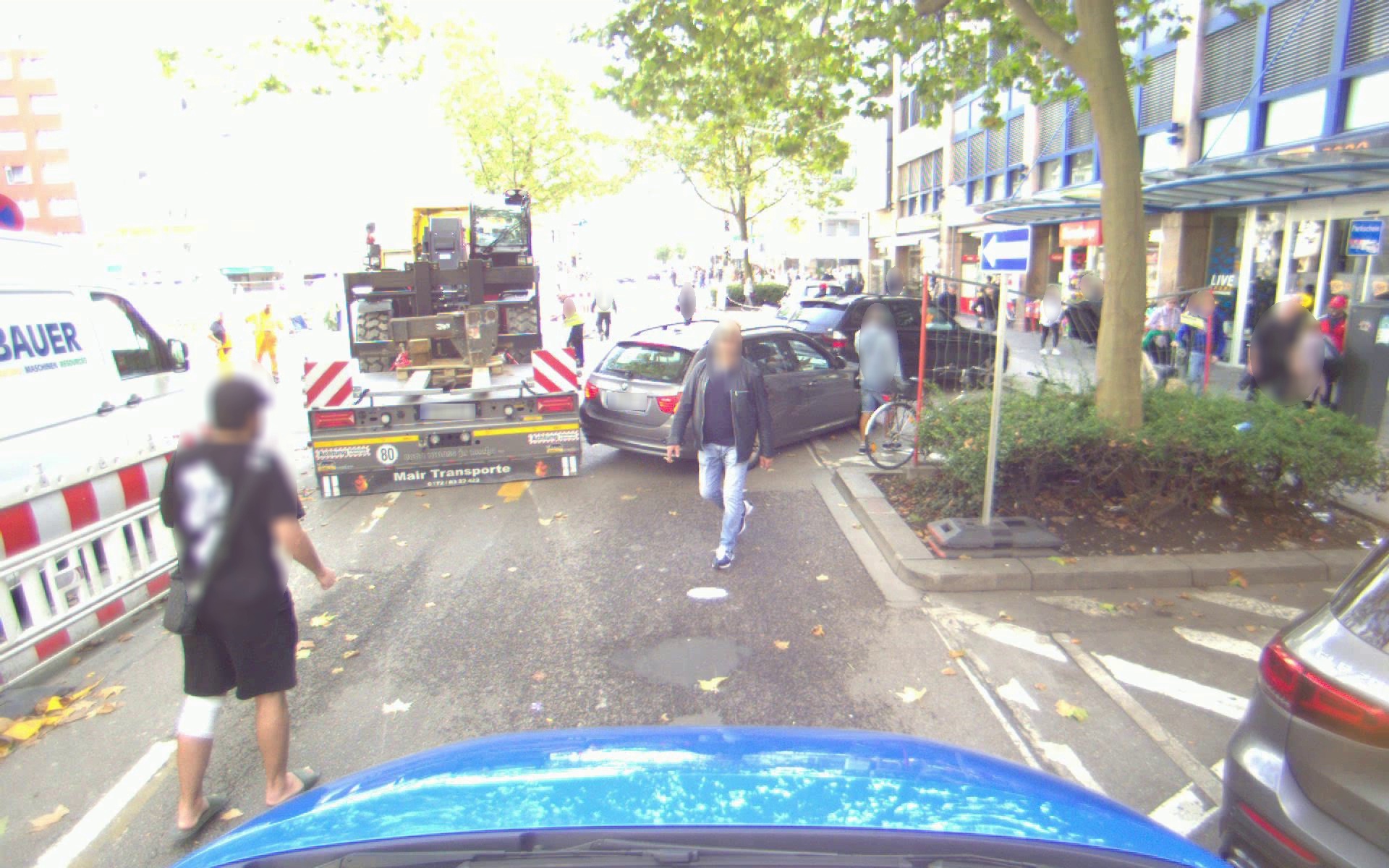}
    \end{subfigure}
    \hfill
    \begin{subfigure}[b]{0.49\linewidth}
        \centering
        \includegraphics[width=\linewidth]{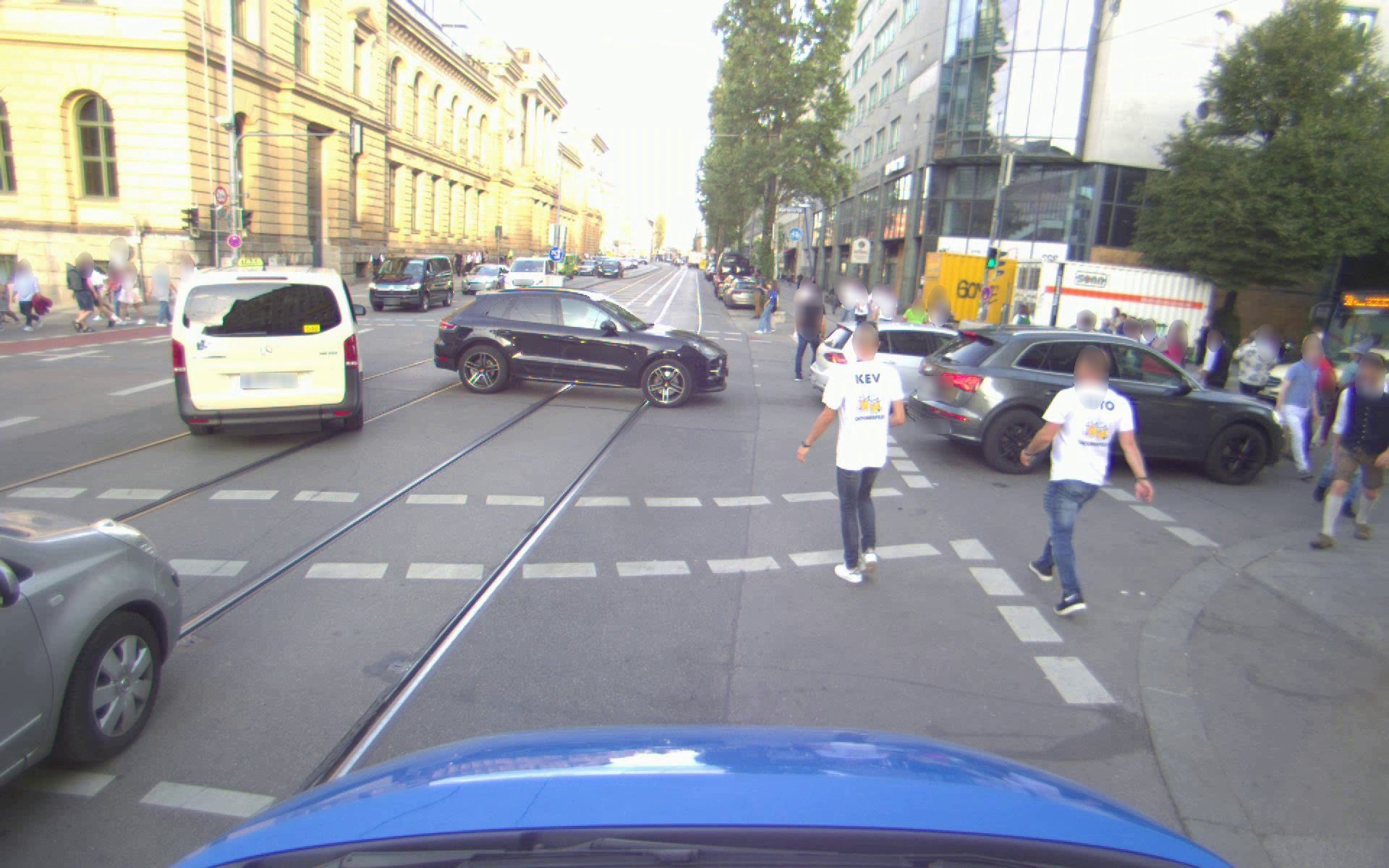}
    \end{subfigure}
    \caption{Examples for miss-classified traffic scene samples, showing a \textit{construction zone} with a parking vehicle on the left and \textit{congested traffic} at a busy intersection on the right.}
    \label{fig:traffic_scene}
\end{figure}

\paragraph{Traffic scene}
When identifying various traffic scenes, Llava-1.6-13 performs best with an F1 score of \SI{52.2}{\percent} and an accuracy of \SI{83.7}{\percent}. However, as shown in \cref{fig:spider}, GPT4-Vision and Composer-HD achieve significantly higher accuracies. This category, like road and weather conditions, often lacks sufficient cues in a single frame, as illustrated in \cref{fig:traffic_scene}. Here, the left image shows a construction scene, and the right depicts congested traffic, both frequently misclassified as traffic accidents. In the first case, it cannot directly be determined whether the gray vehicle in front is moving or stopped, while in the second, objects are clearly in motion. Hence, additional frames would provide crucial context to accurately interpret these scenes.

\begin{figure}[h!]
    \centering
    \begin{subfigure}[b]{0.49\linewidth}
        \centering
        \includegraphics[width=\linewidth]{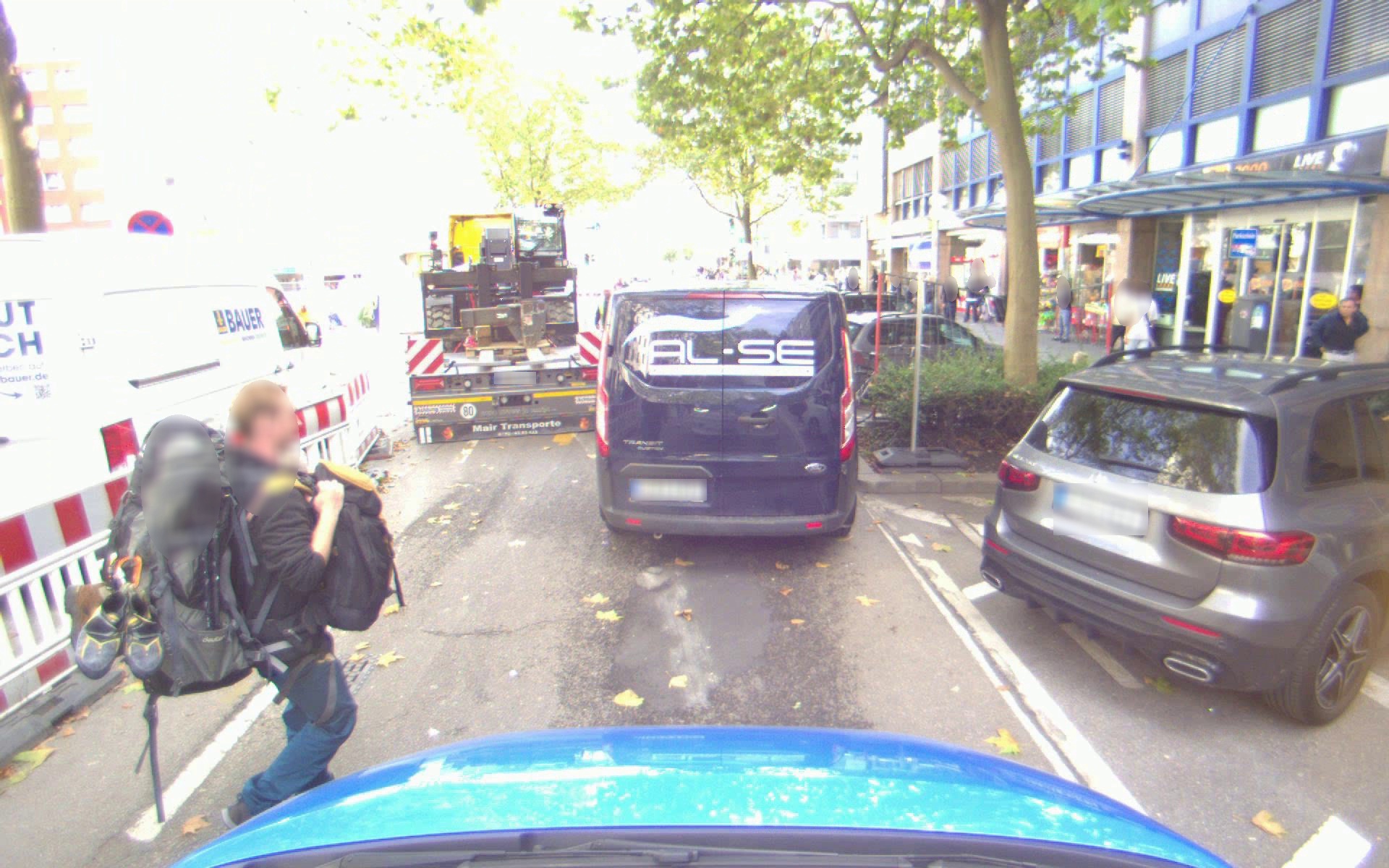}
    \end{subfigure}
    \begin{subfigure}[b]{0.49\linewidth}
        \centering
        \includegraphics[width=\linewidth]{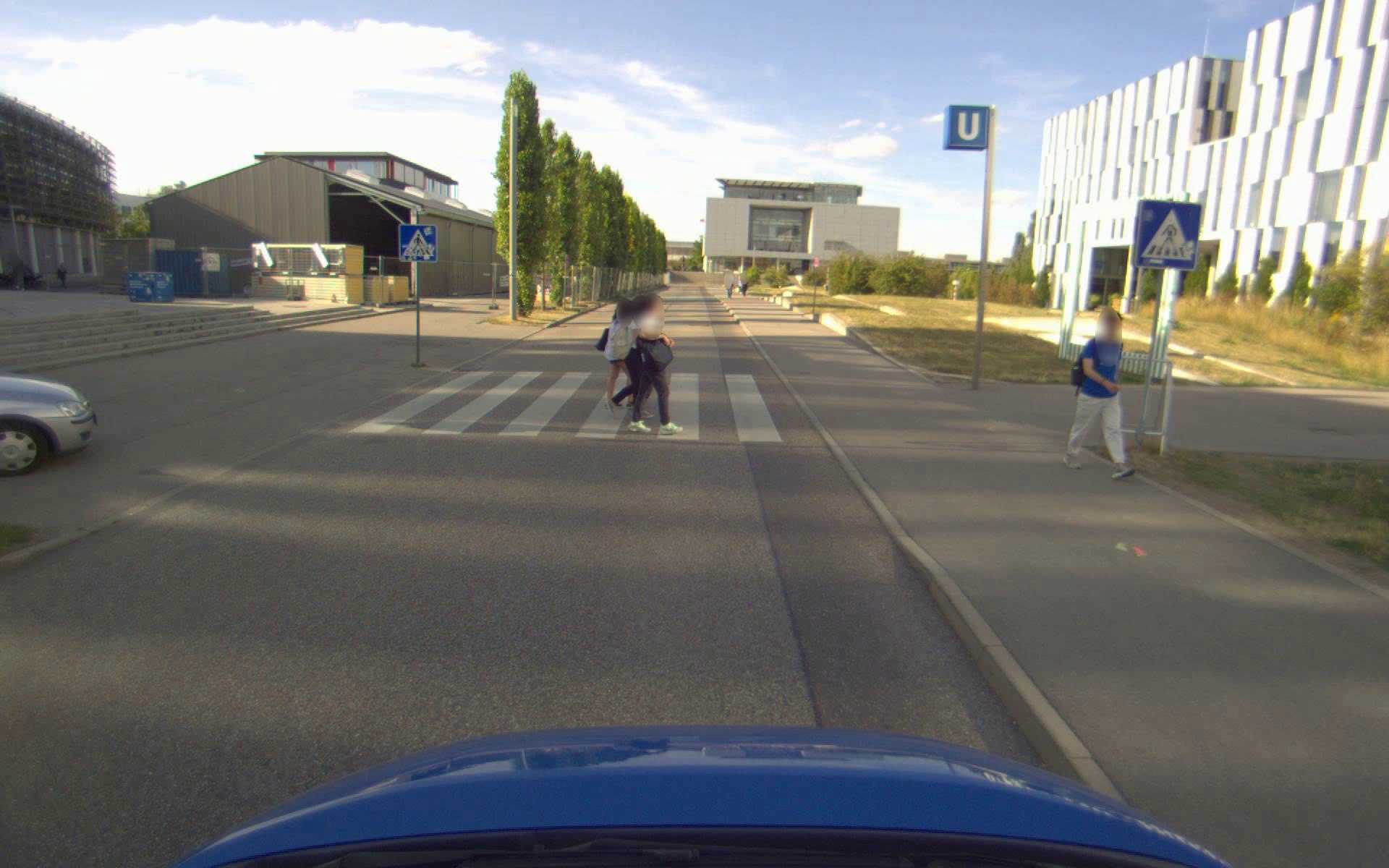}
    \end{subfigure}
    \caption{Example scenarios representing a \textit{stopped} vehicle maneuver on the left and a \textit{moving forward} case on the right.}
    \label{fig:vehicle_maneuver}
\end{figure}

\paragraph{Vehicle maneuver}
As with the previous category, a continuous sequence of frames would be ideal for accurately determining the ego vehicle's precise maneuver. Nonetheless, Llava-1.6-34 and GPT-4 Vision achieve promising accuracies of approximately \SI{74}{\percent}. However, this is largely due to the dominance of \textit{moving forward} scenes, resulting in a significantly lower best F1 score of \SI{35.4}{\percent}. The challenge is illustrated in \cref{fig:vehicle_maneuver}, showing a \textit{stopped} scenario on the left and a \textit{moving forward} one on the right. In the first case, the vehicle is clearly stopped as a pedestrian crosses directly in front of it. In the second case, the situation is less clear since pedestrians are also crossing the road. Regardless, the vehicle continued to roll forward in this case given the noticeable distance to the crosswalk. Both cases were often miss-classified, suggesting \textit{moving forward} for the first and \textit{stopped} for the second.

\begin{figure}[h!]
    \centering
    \begin{subfigure}[b]{0.49\linewidth}
        \centering
        \includegraphics[width=\linewidth]{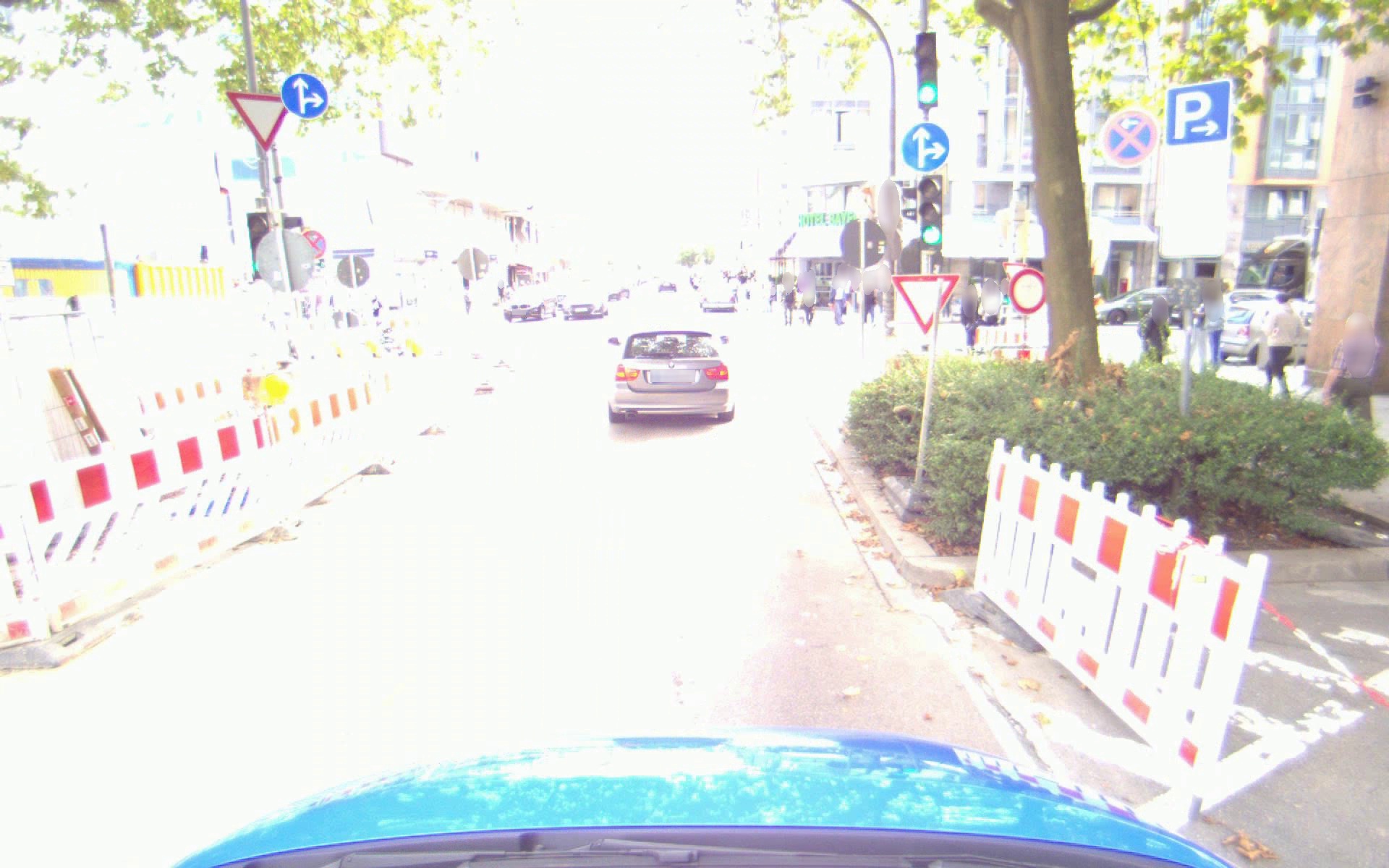}
    \end{subfigure}
    \begin{subfigure}[b]{0.49\linewidth}
        \centering
        \includegraphics[width=\linewidth]{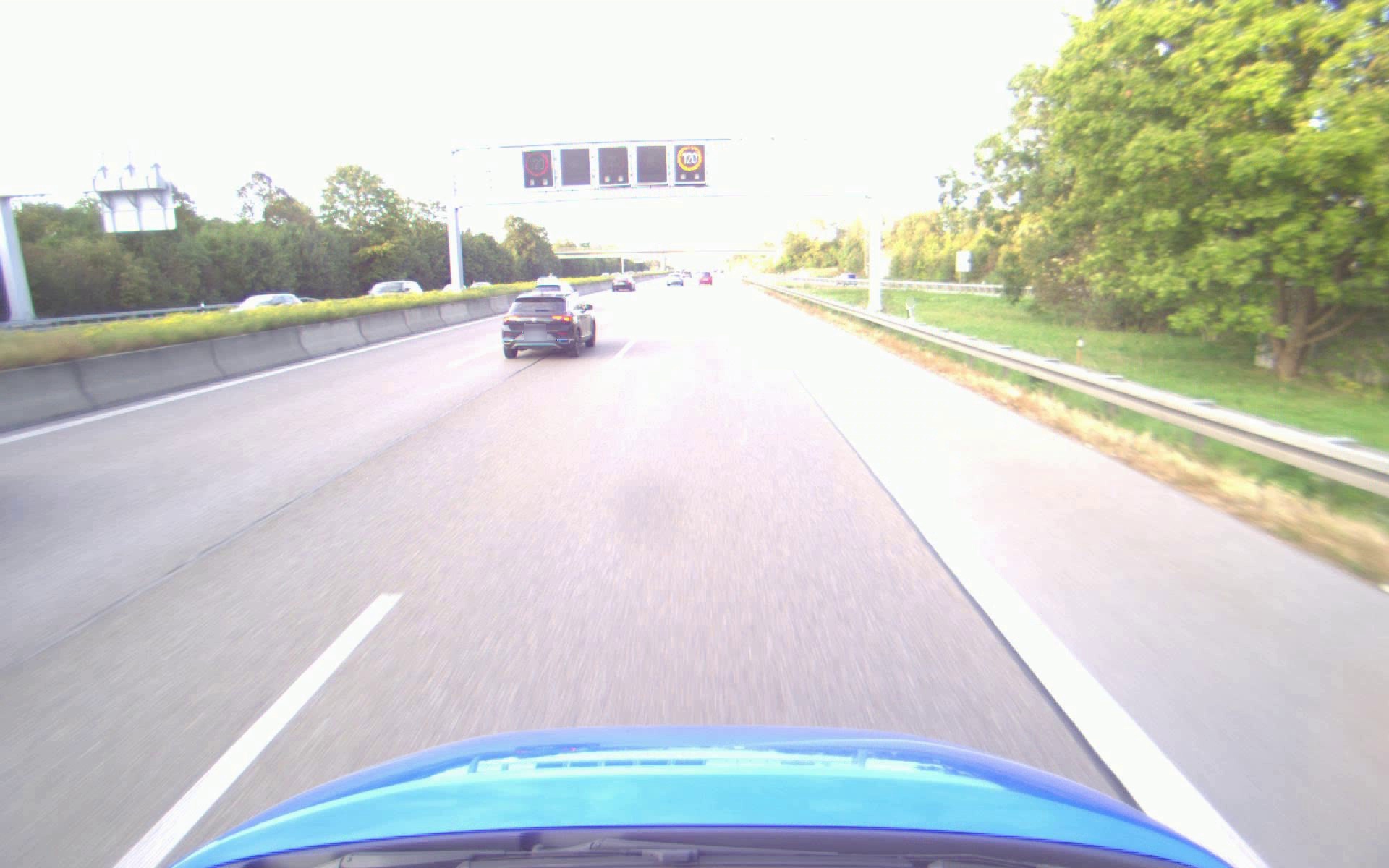}
    \end{subfigure}
    \caption{Samples from the vision-impairing brightness category. Both were classified as too bright, despite manual annotations indicating otherwise as key details remain visible.}
    \label{fig:brightness}
\end{figure}

\paragraph{Vision impairing brightness}
Glare caused by sunlight or reflections can severely impair visibility, sometimes rendering the camera unable to capture the surroundings. While defining an image as "too bright" is subjective, Composer-HD-336 achieves a high F1 score of \SI{81.6}{\percent} and accuracy of \SI{88.7}{\percent} as seen in \cref{tab:bestf1}. In contrast, only GPT-4 Vision performs comparably, with other models showing a significant performance gap as indicated in \cref{fig:spider}. \Cref{fig:brightness} highlights ambiguous classifications, where scenes annotated with no VIB were misclassified by Composer-HD as too bright, despite key details still being visible. This underscores the importance of fine-tuned prompts, as models often interpret visibility in terms of obstacle occlusions rather than brightness and glare.

\paragraph{Street configuration}
The last category presented encompasses the street configuration. Here, Deepseek scores best with an F1 score of \SI{57.0}{\percent} and an accuracy of \SI{63.0}{\percent}. As shown in \cref{fig:spider}, the next-best model, GPT-4 Vision, achieves an F1 score of roughly \SI{50}{\percent}. In terms of accuracy, all models perform similarly well with values around \SI{60}{\percent}. While street configurations can often be inferred from a single frame using cues like road width, lane markings, street signs, or parked vehicle orientation, some cases require additional context. As with traffic scenes or vehicle maneuvers, incorporating extra frames could improve classification performance.

\subsection{BDD100k}
In addition to our in-house dataset, we compare the performance of selected models to traditional closed-vocabulary methods on the BDD100k dataset. We select the best-performing models in the categories "weather" and "urban environment"\footnote{\href{https://github.com/SysCV/bdd100k-models/tree/main/tagging}{https://github.com/SysCV/bdd100k-models/tree/main/tagging}}. ResNet-18 and ResNet-50, fine-tuned on BDD100k, are the top-performing models for this dataset, with \cref{tab:bdd100k} evaluating their overall performance on the BDD100k validation set. Here, ResNets lead in accuracy, outperforming the best LVLMs by \SI{16}{\percent} in the environment and \SI{2}{\percent} in the weather category. However, LVLMs dominate in F1 scores, exceeding both ResNets by \SI{33}{\percent} in the environment category and \SI{5}{\percent} in weather. Causes can be found for nighttime scenarios, where glare complicates classification, and daytime annotations, as they often conflict with our definitions. For example, BDD100k labels scenes as \textit{snowy} when snow is on the ground, while we reserve it for falling snow. Additionally, \textit{overcast} and nighttime conditions are frequently marked as \textit{undefined}. In the urban environment category, accuracy gaps between fine-tuned and general-purpose models are generally smaller. 

\begin{table}[]
    \centering
    \begin{tabular}{@{}lllll@{}}
  
  \toprule
  \multicolumn{1}{c}{}& \multicolumn{2}{c}{Accuracy[\%]}& \multicolumn{2}{c}{F1[\%]}\\
 
    Model & Env & Weather  & Env & Weather \\
     \midrule
    LLaVA-1.6-34 (34B) & 63.3 & 75.1 & \textbf{77.8}&66.2\\
    LLaVA-1.6-13 (13B) & 63.8 &72.3&77.5&69.7\\
    LLaVA-1.6-7 (7B)&51.4& 72.9&77.2&55.3\\
    LLaVA-1.5 (13B) & 48.9 &71.9&77.6&51.7\\
    CogVLM (18B) &61.4 &71.5&77.8&67.7\\
    CogAgent-VQA (18B) &36.5 &71.9&77.8&34.2\\
    Composer-VL (8B) &58.2&72.9&77.5&58.6\\
    Composer-HD (8B) & 50.5 &66.8&74.8&49.4\\
    Deepseek (8B) & 65.0 &71.0&75.3&\textbf{71.8}\\
    ResNet-18  &81.1 &77.1&43.7&66.0\\
    ResNet-50 &\textbf{81.2} &\textbf{77.1}&43.5&65.7\\
    \hline
\end{tabular}
    \caption{Mean Accuracy and F1 score for different models on the BDD100k dataset, listed for the weather and urban environment category.}
    \label{tab:bdd100k}
\end{table}
\section{Discussion}
As seen in the previous section, the performance varied significantly across categories, driven by differences in label ambiguity and temporal context requirements. For instance, in the weather category, it is often unclear which label is most appropriate, while categories like "person" have little ambiguity in determining whether or not a person is present. Temporal context also impacts performance as categories such as street configuration, vehicle maneuver, and traffic scene often require multiple frames for accurate labeling. Additionally, category interdependencies play a crucial role, as seen in the VIB category. When labeled as \textit{yes}, it implies no relevant information can be extracted, rendering the outputs of other categories meaningless. Therefore, future research could explore how these interdependencies affect overall model performance metrics.

\begin{figure}[]
    \centering
    \begin{subfigure}[b]{0.49\linewidth}
        \centering
        \includegraphics[width=\linewidth]{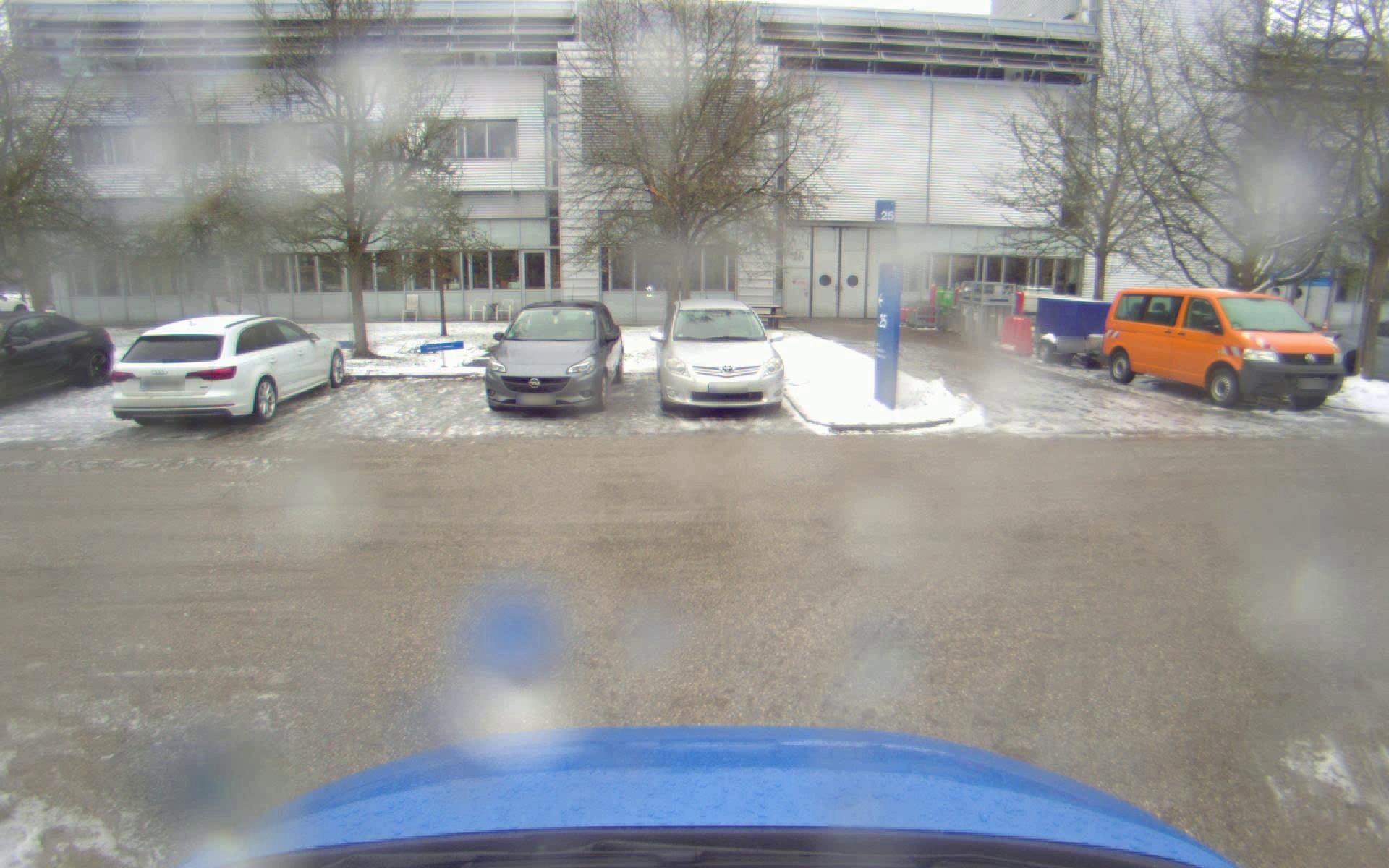}
    \end{subfigure}
    \begin{subfigure}[b]{0.49\linewidth}
        \centering
        \includegraphics[width=\linewidth]{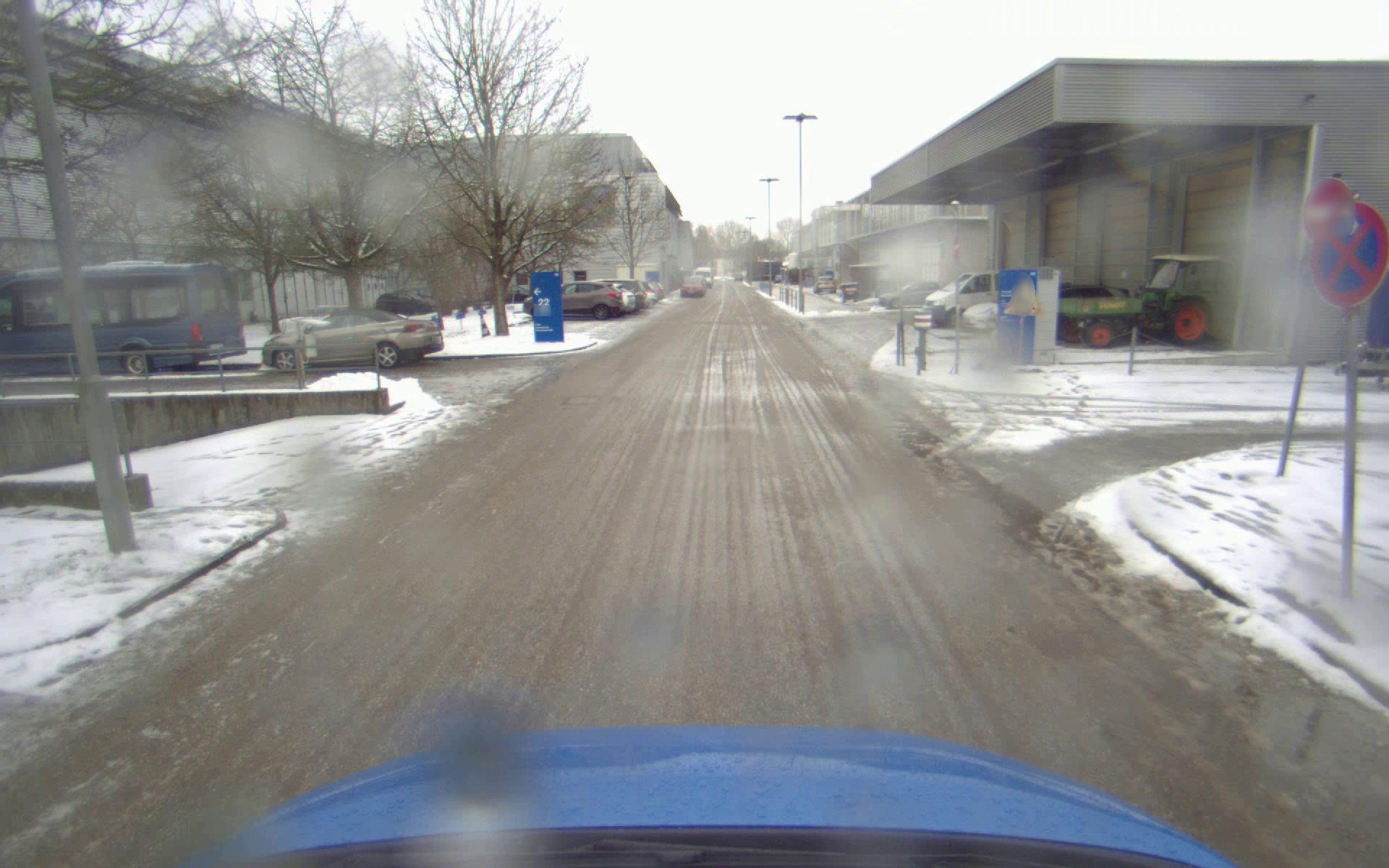}
    \end{subfigure}
    \caption{Samples in the road condition category, highlighting that both \textit{snowy} an \textit{muddy} labels are reasonable depending on ones defintion.}
    \label{fig:road_condition}
\end{figure}

While LLaVA-1.5 stands out as the best model based on its strong F1 score and mean accuracy, these metrics alone can be misleading, particularly with smaller datasets. Accuracy may overestimate performance, while a low F1 score might not accurately reflect poor performance when labels are sparse or edge cases dominate. This highlights the importance of qualitative analysis. For instance, the road condition category exhibits the highest ambiguity, as illustrated in \cref{fig:road_condition}. The two examples were both labeled as \textit{muddy}, while Composer-HD missclassified the one on the right with \textit{snowy}. Arguably, both labels could be applied for both cases when a clear definition is lacking.

Overall, the Composer-HD model demonstrates the best qualitative performance across several categories. However, its lower F1 scores in reasoning tasks stems from ambiguous ground truth labels in categories like road condition and time of day, where challenging tags had fewer than ten samples. In contrast, GPT-4 excels in reasoning tasks, showing strengths in interpreting vulnerable road users and handling complex scenarios, but struggles with traffic signs. Using Composer-HD with a 336px input resolution reduces its performance further, as the tags for categories reliant on specific image details, such as the number of lanes, cannot be infered properly anymore. To improve performance further, the models' input should be adjusted to handle several frames, providing enhanced contextual understanding through temporal reasoning.

The quantitative comparison on the BDD100K dataset reveals a divergence between accuracy and F1 score: While the accuracy favors classical CNNs, the F1 scores are higher for LVLMs. This disparity may stem from the BDD100K data distribution \cite{yu2020bdd100kdiversedrivingdataset}, where weather tags are evenly distributed, but environment tags are skewed towards fewer classes. Hence, since the selected LVLMs were not trained on BDD100K distribution, their strong F1 performance highlights their generalization capabilities. In addition, LVLMs also excel in inference and don't require retraining for new categories, aligning with our goal of cross-domain generalization. Overall, based on the results provided, they offer a more versatile and efficient solution for the presented use-case compared to task-specific classifiers, with the added benefit of reasoning about traffic states or movement-related tasks.

\subsection{Inference Time}
As our goal is to classify scenes in an existing dataset and the dataset is relatively small, the inference time is not a critical factor for us. For larger datasets and time restricted computational resources, this factor might become important. Therefore, we measured the inference time per forward pass. As prompt we used “Tell me about the image.” and we restricted the number of output tokens to 10. 
We use one A-100-80GB GPU for the experiments. We discard GPT-4 at this point because it's API calls can be parallelized. Composer-HD has a significantly larger inference time with almost 15 seconds per run compared to the LLaVA models, almost 10x slower than the slowest Llava. This will have an impact with larger datasets and more categories. The inference times shown indicate upper bounds as the output is generally shorter than ten tokens.

\begin{table}[tb]
  \caption{Processing times for a single image.
  }
  \label{tab:inference}
  \centering
  \begin{tabular}{@{}lll@{}}
    \toprule
    Model Name &Latency[s]\\
    \midrule
    LLaVA-1.6-34(34B) & 1.549\\
    LLaVA-1.6-13 (13B) & 0.77\\
    LLaVA-1.6-7 (7B)&0.514\\
    LLaVA-1.5 (13B) & \textbf{0.422}\\
    CogVLM (18B) &5.984\\
    CogAgent-VQA (18B) &4.922\\
    Composer-HD (8B) & 14.953\\
    Composer-HD-336px (8B) & 4.785\\
    Deepseek (8B) & 0.826\\
  \bottomrule
  \end{tabular}
\end{table}

\section{Conclusion}
In this work, we evaluated different LVLMs' classification performance on an autonomous driving
dataset. The models were selected based on their performance on relevant VQA datasets. We
saw that the best models performed strongly on various detection tasks. Additionally, the models
showed good reasoning capabilities. Their ability to classify traffic scenarios and conclude the
appropriate vehicle maneuver from single images was especially remarkable. All models had
difficulties classifying the street configuration, which is the only category for which the quality of
predictions is insufficient for integration in this project. The land-use category predictions were
inaccurate but provided a good understanding of the vehicle’s surroundings. If there is enough
time to execute the initial prediction run with the Composer-HD model, this will result in optimal
classification performance using a single model. The LLaVA-1.5 model provides an overall solid
performance in all categories and is a good alternative with a much smaller inference time. GPT-4 was the best-performing model in the more complex reasoning task and is a good choice for
the corresponding categories if the financial resources are available. 

We have shown that LVLMs can be used to automatically classify scenes in autonomous driving datasets. The quality of classifications might be improved by additionally using other modalities like 3D point clouds provided by LiDAR, GPS, or IMU data. For the categories vehicle maneuver, street configuration and traffic scenario improvements can likely be obtained by using multiple frames as model input. Also, the size of our dataset is not sufficient for extensive evaluation of rare scenarios: We have seen that in edge cases where multiple labels can fit, the comparison across models might profit from having more labeled data available. 

In conclusion, this study presents promising findings, demonstrating that LVLMs can play a crucial role in the automated characterization of traffic scenarios. By improving upon existing CNN-based methods, LVLMs offer a more scalable solution to the scene understanding problem. These advancements pave the way for further research into leveraging LVLMs for more efficient and comprehensive analysis of complex traffic environments.

{\small
\bibliographystyle{ieee_fullname}
\bibliography{PaperForReview}
}

\end{document}